\documentclass[10pt,twocolumn,letterpaper]{article}

\usepackage{cvpr}              

\usepackage{graphicx}
\usepackage{amsmath}
\usepackage{amssymb}
\usepackage{booktabs}
\usepackage{xcolor}
\usepackage{tabularx}
\usepackage{multirow}
\usepackage[accsupp]{axessibility}  

\usepackage[pagebackref,breaklinks,colorlinks]{hyperref}

\usepackage[capitalize]{cleveref}
\crefname{section}{Sec.}{Secs.}
\Crefname{section}{Section}{Sections}
\Crefname{table}{Table}{Tables}
\crefname{table}{Tab.}{Tabs.}

\def\cvprPaperID{3547} 
\def\confName{CVPR}
\def\confYear{2022}

\begin{document}

\title{CDGNet: Class Distribution Guided Network for Human Parsing}
\author{Kunliang Liu$^{1,2}$, Ouk Choi$^{3}$, Jianming Wang$^{2*}$, and Wonjun Hwang$^{1*}$\\
$^{1}$Ajou University, Korea,  $^{2}$Tiangong University, China, $^{3}$Incheon National University, Korea\\
{\tt\small tjpulkl@ajou.ac.kr, ouk.choi@inu.ac.kr, wangjianming@tiangong.edu.cn, wjhwang@ajou.ac.kr}
}
\maketitle

\let\thefootnote\relax\footnote{* Equal contribution corresponding author}

\begin{abstract}
The objective of human parsing is to partition a human in an image into constituent parts. This task involves labeling each pixel of the human image according to the classes. Since the human body comprises hierarchically structured parts, each body part of an image can have its sole position distribution characteristic. Probably, a human head is less likely to be under the feet, and arms are more likely to be near the torso. Inspired by this observation, we make instance class distributions by accumulating the original human parsing label in the horizontal and vertical directions, which can be utilized as supervision signals. Using these horizontal and vertical class distribution labels, the network is guided to exploit the intrinsic position distribution of each class. We combine two guided features to form a spatial guidance map, which is then superimposed onto the baseline network by multiplication and concatenation to distinguish the human parts precisely. We conducted extensive experiments to demonstrate the effectiveness and superiority of our method on three well-known benchmarks: LIP, ATR, and CIHP databases.
$^{\dagger}$
\footnote{$^{\dagger}$Our code is available at https://github.com/tjpulkl/CDGNet.}
\end{abstract}

\section{Introduction}
\label{sec:intro}
Human parsing involves segmenting human bodies into constituent parts, such as the head, body parts, and clothing items. It belongs to the field of scene parsing, where per-pixel categorization is performed for a given image. Human parsing is highly challenging owing to the complex textures and styles of clothes, deformable human poses, and scale diversity of different semantic parts. Human parsing enables fine-grained semantic segmentation, benefits human understanding, and supports human-centric applications.

\begin{figure}[t]
\centering
\includegraphics[width=1.0\linewidth]{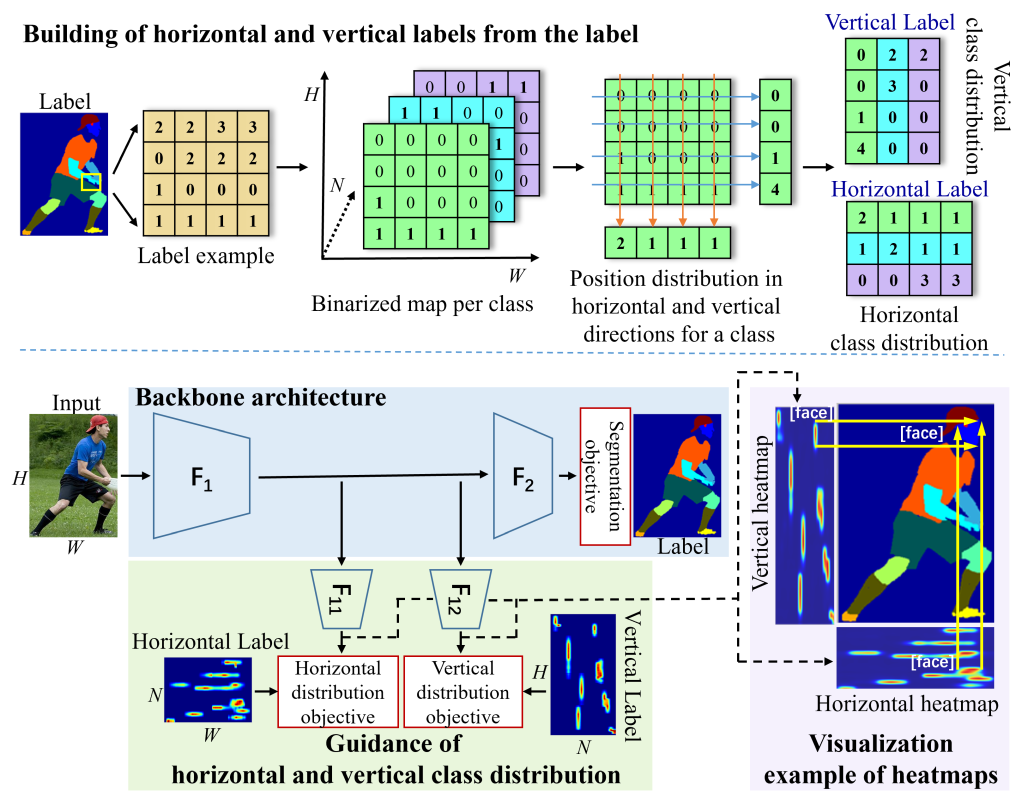}
\caption{\textbf{Guidance of class distribution}. (Top) We build the novel horizontal and vertical class distribution labels from the original label to simplify the 2D spatial human parsing problem with 1D positional labels in the horizontal and vertical directions. (Bottom-left) We directly teach the backbone network with our horizontal and vertical class distribution labels as a positional guidance signal of human parts. (Bottom-right) For instance, the horizontal and vertical face positions are correctly estimated as observed in the visualization heatmap examples.
}
\label{fig:FusedGTImg}
\vspace{-0.6cm}
\end{figure} 

Since the emergence of fully convolutional networks (FCNs)~\cite{Jonathan2015FCN-13}, various studies have developed several solutions from different perspectives to boost the performance of this dense prediction task. Several previous methods~\cite{Liang_Atrous_seg,Fu2019ACNet-2,Hou2020StripPooling-3,Fu2019DANet-7,Huang2019CCNet-9,Yuan2020OCNet-10,Yuan2018Ocnet-11,Zhao2019PPM-17,Zhao2020CPSS-18,Ruan2019CE2P-21,ISNet2021ICCV} using both spatial and semantic contexts have achieved great successes in scene segmentation; however, human parsing has unique characteristics that make it much harder than the conventional pixel-level segmentation tasks. 
Specifically, the human body comprises symmetrical and structural parts. 
The left and right arms, and the left and right shoes, have a similar appearance as well as a similar vertical position but must be separated into different classes. Naturally, in this case, we utilize the horizontal position difference between the left and right human parts. Furthermore, the scarf and glasses are too small parts to be segmented; however, most of them are on the face or the neck, not on the legs, and we can obtain better performance when using this structural knowledge successfully. 
Several works took advantage of human characteristics for achieving better performance in human parsing. 
Ji et al.~\cite{Ji2020LSNT-22} designed a novel semantic neural tree to encode the physiological structure of the human body and achieved competitive results. Wang et al.~\cite{Wang2020HTypePRR-20,Wang2019LCNinfo-25} exploited deep graph networks and hierarchical human structures to capture the relation information of human parts and obtained better performances. These mechanisms involve designing a complex semantic tree or message-passing network that leads to heavy computing complexity while improving performance. Moreover, building different tree structures for diverse datasets limits the deployment of these methods. Zhang et al.~\cite{Zhang2020CorrPM-8} utilized human pose and non-local mechanism to achieve good performance, but it required accurate human pose information in advance.
Consequently, most recent works have attempted to solve complex human parsing problems by utilizing complex modules such as graph network~\cite{Wang2020HTypePRR-20} and human pose estimator~\cite{Zhang2020CorrPM-8}.

In this paper, we attempt to solve the human parsing problem following the \textit{divide and conquer} strategy. We simplify the human structure information complexly represented in 2D space into horizontal and vertical 1D position information with the corresponding classes. To this end we propose a novel class distribution guided network (CDGNet) taught by human categorical positional knowledge represented in the horizontal and vertical directions. As shown at the top of \cref{fig:FusedGTImg}, we first build the horizontal and vertical class distributions as new supervision signals by accumulating the horizontal-wise and vertical-wise binarized map for each class from the original label. We squeeze the baseline feature in the horizontal and vertical directions and then teach them using the corresponding class distribution labels, as described in the bottom of \cref{fig:FusedGTImg}. For example, the face candidate region is successfully represented by the face class distributions at the right bottom of \cref{fig:FusedGTImg}. Note that we can use all the classes without selecting a few specific classes for learning the model because we have simplified the complex human parsing problem. Finally, in order to accomplish the precise human paring results, we merge these guided features and then superimpose them on the backbone network features.

The major contributions of our work are summarized as follows: 

   \noindent$\bullet$ We simplify the complex spatial human parsing problem into the horizontal and vertical positions of human parts individually. Accordingly, we build the class distribution labels in the horizontal and vertical directions as new supervision signals from the original label of human parsing.\\
    $\bullet$ Using these class distribution labels, we propose the CDGNet that guides the backbone network toward exploiting the intrinsic position distribution of human parts.\\
    $\bullet$ We verify the significant performance improvement gained by the proposed method through extensive experiments on LIP, ATR, and CIHP benchmarks. \\

\section{Related Works}
\textbf{Semantic Segmentation:} 
Human parsing is a fine-grained semantic segmentation task in which all pixels of the human image are labeled. The method utilized in human parsing is similar to that used for semantic segmentation, which predicts each pixel labeling in the scene. Earlier scene parsing studies mainly focused on the spatial scale of contexts~\cite{Huang2019SAN,Chen2018DPC,Liang_Atrous_seg,Chen2018V3,Badrinarayanan2017Segnet-14,olaf2015UNet-15,Chen2017Deeplab-1624,Zhao2019PPM-17,Hou2020StripPooling-3,DualGraphConv4,wang2018Nonlocal-5,wang2019AsyNonlocal-6,Fu2019DANet-7,Zhang2020CorrPM-8,Huang2019CCNet-9,MCIBISS2021ICCV,Chen2016Attentiontoscale,Fu2019ACNet-2,Equivarian2020,Huang2019ISSA-27}.
However, owing to the limitations of the structure of convolutional layers, the spatial context provides insufficient contextual information. Methods such as OCR~\cite{Yuan2020OCNet-10} and ACFNet~\cite{Zhang2019ACFNet-12} group the pixels into a set of regions, augment the pixel representations by region representations to which the pixel belongs, and gain competitive performance on various challenging semantic segmentation benchmarks; however, these methods did not intentionally consider the spatial distribution of each category, confining the ability to capture the distribution of different classes, and cannot utilize the distribution rule to benefit parsing efficiently.

\textbf{Human Parsing:} 
Several deep learning-based methods have achieved significant improvements in human parsing. Ruan et al.~\cite{Ruan2019CE2P-21} designed a CE2P network based on ResNet-101 that fully leveraged feature resolution, global spatial context information, and edge detail. It achieved its first place in LIP Challenge in 2019. Given the label of a pixel is the category of the object to which the pixel belongs, Yuan et al.~\cite{Yuan2018Ocnet-11} implemented an object-contextual representation approach for semantic segmentation and achieved an even higher performance on LIP. Other studies have combined additional human prior information for human parsing. For instance, Wang et al. \cite{Wang2020HTypePRR-20} assembled the compositional hierarchy of human bodies for efficient and complete human parsing, and Ji et al.~\cite{Ji2020LSNT-22} exploited the intrinsic physiological structure of the human body by designing a novel semantic neural tree for human parsing. Utilizing grammar rules in a cascaded and parallel manner, Zhang et al.~\cite{Zhang2020BGNet-23} employed the inherent hierarchical structure of the human body and the relationship of different human parts to achieve impressive human parsing results. Zhang et al.\cite{Zhang2020CorrPM-8} combined human semantic boundaries and keypoint locations to improve human parsing. These methods either depend on the prior human hierarchical structure, or the precise human pose, which makes it difficult to guarantee generality if there are multiple persons or if human parts are covered by unexpected occlusions.

\section{Class Distribution Approach}
In this section, we propose an efficient method that utilizes human part class distributions for human parsing without additional heavy networks, e.g., human pose detection~\cite{Zhang2020CorrPM-8} and complex assumption~\cite{Wang2020HTypePRR-20}. Our method is catalyzed by attention-based methods such as SENet~\cite{SENet}, CBAM~\cite{CBAM} and HANet~\cite{Choi2020HDANet-19}. SENet and CBAM captured the global context of the entire image for general purposes. HANet considered solely a height-driven attention map because it focused on urban scene images.

\begin{figure}[t]
\centering
\includegraphics[width=0.7\linewidth]{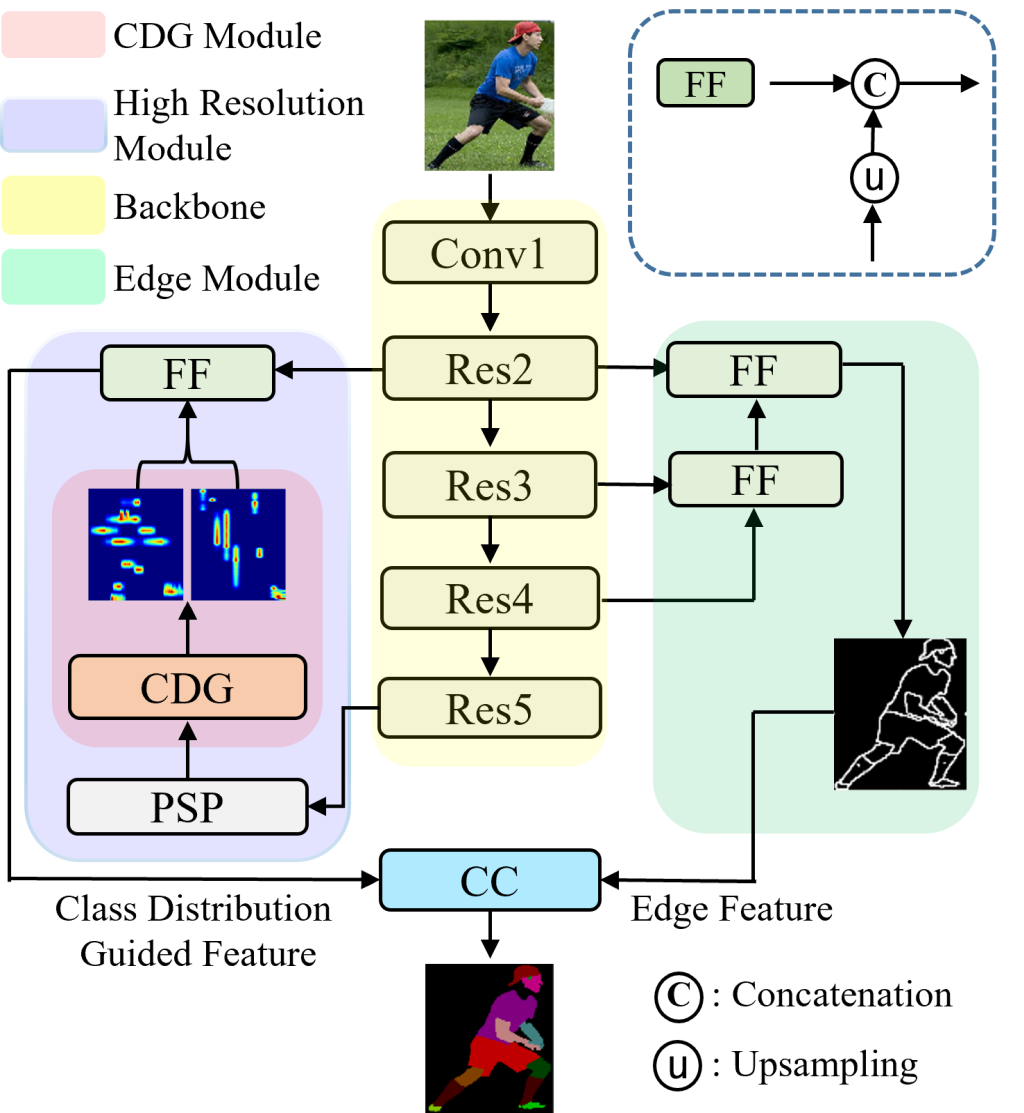}
\caption{\textbf{Overview of our network.} \textbf{FF}: feature fusing. \textbf{PSP}: pyramid spatial pooling. \textbf{CC}: concatenation and convolution of feature map. \textbf{CDG}: our class distribution guidance module.  
}
\label{fig:backbone}
\vspace{-0.6cm}
\end{figure}

In this paper, rather than employing the attention methods, we extend the original human parsing label to another supervision signals related to classes and one directional positions. These generated signals play a pivotal role in guiding the network to locate the human parts efficiently. Note that the attention mechanism has not used the supervision signals for enhancing the feature representation.
Because human bodies are hierarchically structured as described in~\cite{Wang2020HTypePRR-20}, which means that it has significantly strict spatial distribution features, the distinct parts of a human body have different manifest distributions in the vertical and horizontal dimensions. We propose a class distribution guided network that predicts the class distribution in the vertical and horizontal dimensions under the supervision of class distribution loss. The predicted distribution characteristic of the human image is then fully employed to enhance the feature representation for human modeling. The overall proposed procedure is summarized in~\cref{fig:backbone}.

\begin{figure*}[ht]
\centering
\includegraphics[width=1.0\linewidth]{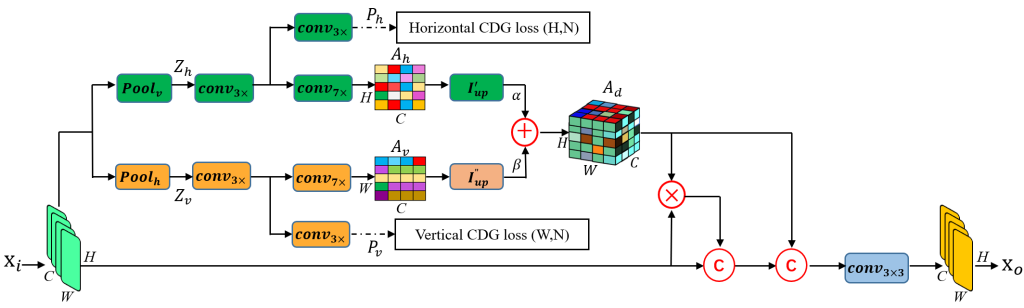}
\caption{\textbf{Architecture of the proposed CDGNet.} $\boldsymbol{X}_i$: input feature, $\boldsymbol{Z}_h$ and $\boldsymbol{Z}_v$: horizontal and vertical features, $\boldsymbol{A}_h$ and $\boldsymbol{A}_v$: horizontal and vertical guided features, $\boldsymbol{A}_d$: spatial guidance map, $\boldsymbol{X}_o$: CDG feature, $Pool_v$: pooling along vertical direction to obtain horizontal distribution, $Pool_h$: pooling along horizontal direction to obtain vertical distribution, $conv_{3\times 3}$: $3\times 3$ kernel-based 2D convolution, $conv_{3\times}$, and $conv_{7\times}$: 1D convolution with kernel size $3$ and $7$, $I^{'}_{up}$ and $I^{''}_{up}$: upsampling horizontal and vertical guided features, $N$: the category number of human part, $H$, $W$ and $C$: height, width and channel number of the input feature map, $\alpha$ and $\beta$: coefficients of horizontal and vertical guided features when aggregating them. ${\color{red}\bigoplus}$: 
element-wise summation, ${\color{red}\bigotimes}$: element-wise multiplication of the matrix, and {\color{red}{\textcircled{$\textbf{c}$}}}: feature concatenation. 
}
\label{fig:hwdnetwork}
\vspace{-0.4cm}
\end{figure*}

\subsection{Horizontal and Vertical Class Distribution}
In human parsing, images with per-pixel human-part labels are given as a training dataset. 
From the labels, we calculate the per-class positional distributions within each image in the horizontal and vertical directions, respectively,
which are called horizontal and vertical class distributions in this paper. 
The class distributions guide the network to learn the distribution context of each category, which forces the network to consider different category spatial distributions under the constraint of the proposed distribution loss. 
Given an input image $\boldsymbol{I}$ and the corresponding label~$\boldsymbol{G}$, we encode each class integer label to one-hot encoding, generating a matrix $\boldsymbol{M}$ of $H\times W\times N$ size, where $H$ and $W$ are the input image height and width, respectively, and $N$ is the class number. As the top part of \cref{fig:FusedGTImg} indicates, in each channel, the number of `$\boldsymbol{1}$' is counted along the horizontal and vertical directions, generating the vertical and horizontal class distributions. Therefore, the two distributions, $\boldsymbol{G_{D_h}}$ of size $W\times N$ and $\boldsymbol{G_{D_v}}$ of size $H\times N$ can be obtained by the following equations; 
\begin{equation}
   \boldsymbol{G_{D_h}}(w,n)=\sum_{i=1}^{H}\boldsymbol{M}_{i,w,n},~
   \boldsymbol{G_{D_v}}(h,n)=\sum_{i=1}^{W}\boldsymbol{M}_{h,i,n},
    \label{eq:Dhc2}
\end{equation}
where $h\in \{1\sim H\}$, $n\in \{1\sim N\}$, $w\in \{1\sim W\}$, $\boldsymbol{G_{D_h}}$ and $\boldsymbol{G_{D_v}}$ are the horizontal and vertical class distributions as new supervision signals for human parsing, respectively.

\subsection{Class Distribution Guided Network}
Our CDGNet produces each class distribution that reveals where and to what extent the class instance exists and correctly guides the feature representation for human parsing. The structure of CDGNet is illustrated in \cref{fig:hwdnetwork}.
When inputting a feature, $X_i$ of $W\times H\times C$ where $C$ is the channel size, we squeeze the input feature in the horizontal and vertical directions, individually, to extract the directional position characteristics. Accordingly, we employ average pooling in orthogonal directions, e.g., vertical average pooling for the horizontal feature $\boldsymbol{Z}_{h}$ and horizontal average pooling for the vertical feature $\boldsymbol{Z}_{v}$. 
After average pooling, one 1D convolution with kernel size $3$ followed by Batch Normalization (BN) layer is leveraged to decrease the channel number, i.e., $C/2$. 
Because the number of classes is generally smaller than the channel size, it is necessary to reduce the channel size by half as a buffer.

To generate the horizontal and vertical class distribution features, we utilize 1D convolution with kernel size $3$ with its channel number, $N$, and we guide these features using the new labels regarding the horizontal and vertical class distributions through the corresponding losses. 
Meanwhile, another 1D convolution with kernel size $7$ and the same channel number with the input feature, $C$, is applied to produce each channel distribution feature in the horizontal and vertical directions, separately. 
The two convolutions are activated by the $sigmoid$ function, instead of the $softmax$ function to produce the salient distribution maps. 
The reason is that multiple classes exist in each row of $\boldsymbol{Z}_{h}$ and $\boldsymbol{Z}_{v}$. 
These series of operations for generating the guided features on the horizontal or vertical dimensions of each channel can be denoted as:
\begin{equation}
    \boldsymbol{A}_{h}^{'}=I_{up}^{'}(\boldsymbol{A}_{h})
    =I_{up}^{'}(\sigma (conv_{7\times}(\delta (conv_{3\times}(\boldsymbol{Z}_{h}))))),
    \label{eq:achw1}
\end{equation}
\begin{equation}
    \boldsymbol{A}_{v}^{'}=I_{up}^{''}(\boldsymbol{A}_{v})
    =I_{up}^{''}(\sigma (conv_{7\times}(\delta (conv_{3\times}(\boldsymbol{Z}_{v}))))),
    \label{eq:achw2}
\end{equation}
where $\sigma$ is a $sigmoid$ function, and $\delta$ is a $ReLU$ function. 
The guided feature $\boldsymbol{A}_{h}$ involves the position-related features for each class in the channel according to the horizontal direction whereas the guided feature $\boldsymbol{A}_{v}$ includes the vertical position-related class features in the channel because they are all guided by the corresponding CDG loss functions. 
Bilinear interpolation operations $\boldsymbol{I}_{up}^{'}$ and $\boldsymbol{I}_{up}^{''}$ are then applied to the corresponding guided features to expand their other dimensions to have the same dimension as the input feature $\boldsymbol{X}_i$. 
The extended guided features $\boldsymbol{A}_{h}^{'}$ and $\boldsymbol{A}_{v}^{'}$ are weighted by coefficients $\alpha$ and $\beta$ and merged by the element-wise sum operation to combine the separately learned models. 
We adopt the element-wise multiplication and concatenation sequentially to properly reflect the spatial guidance map $\boldsymbol{A}_d$ to the input feature. Using this scheme, the input feature $\boldsymbol{X}_i$ can be transformed into a CDG feature $\boldsymbol{X}_o$. 
Finally, a $3\times 3$ kernel-based convolution is adopted to fit the output feature size identical to the input feature size. 
The formula is as follows:
\begin{equation}
\begin{split}
   \boldsymbol{A}_d = \alpha\times \boldsymbol{A}_{h}^{'} + \beta\times \boldsymbol{A}_{v}^{'},
   \end{split}
    \label{eq:xo1}
\end{equation}
where $\alpha$ and $\beta$ are the learnable weight parameters. 

\subsection{Class Distribution Guided Loss}
The human body is non-rigid, and a few human parts can be occluded during vigorous exercise. These facts pose a difficult challenge in improving the accuracy of the human parsing model utilizing only 2D human body labels.
The edge generated from the label~\cite{Ruan2019CE2P-21, Zhang2020CorrPM-8} is used to clarify the boundaries between human parts to solve this problem. If we first know where it is most likely to determine the different human part classes in the feature map, the problem can be alleviated. In this respect, we introduce the CDG loss for teaching the network to model each category distribution, where pixels on each row in vertical and horizontal distribution maps indicate the related category distribution in both directions. 

The predicted class distribution maps are $P_{h}$ of size $W$ and $N$, and $P_{v}$ of size $H$ and $N$ as shown in~\cref{fig:hwdnetwork}. 
Assume that each position in the predicted horizontal class distribution map is $\{{p_{i,j}\in P_{h}, i\in [1,W], j\in [1,N]}\}$, and in the vertical class distribution map is $\{p_{i,j}^{'}\in P_{v}, i\in [1,N], j\in[1,H]\}$. Accordingly, assuming that the corresponding positions of the horizontal class distribution label are $\{{g_{i,j}\in G_{D_h}, i\in [1,W], j\in [1,N]}\}$, and the vertical class distribution label is $\{g_{i,j}^{'}\in G_{D_v},i\in [1,N],j\in [1,H]\}$. The L2 loss between the prediction and label can be denoted as:
\begin{equation}
  l_{h}=\frac{1}{W\times N}\sum_{i=1}^{W}\sum_{j=1}^{N}(p_{i,j} - g_{i,j})^2,
  \vspace{-0.2cm}
\end{equation}
\begin{equation}
  l_{v}=\frac{1}{N\times H}\sum_{i=1}^{N}\sum_{j=1}^{H}(p_{i,j}^{'} - g_{i,j}^{'})^2.
\end{equation}

The complete CDG loss $l_{CDG}$ is denoted as follows;
\begin{equation}
  l_{CDG}=\theta \times l_{h} + \phi \times l_{v},
  \label{eq:loss1}
\end{equation}
where $\theta$ and $\phi$ are the balance weights for the horizontal and vertical CDG losses, respectively. The weights are set as $\theta=1$ and $\phi=1$ for the simplicity in this paper.

\subsection{Training Objectives}
In this paper, CE2P~\cite{Ruan2019CE2P-21} was used as the baseline network. The CE2P output includes two parsing results and one edge prediction. Thus, the total loss of CE2P can be denoted as:
\begin{equation}
  L_{baseline}=L_{parsing}+L_{edge}+L_{edge-parsing},
  \label{eq:loss2}
\end{equation}
where $L_{edge}$ is the weighted cross-entropy loss between the predicted edge map generated using the edge module and edge label map. Here, $L_{parsing}$ is cross-entropy loss between the parsing map of the high-resolution module and human parsing label, and $L_{edge-parsing}$ is another cross-entropy loss between the parsing label and the final parsing result, after combining the feature maps of edge perceiving module. Our method appends after the pyramid pooling module, as shown in \cref{fig:backbone}. Thus, the final loss function can be formulated using equations~(\ref{eq:loss1}) and (\ref{eq:loss2}) as follows:
\begin{equation}
  L=\tau\times L_{baseline}+\gamma\times l_{CDG},
\end{equation}
where $\tau$, and $\gamma$ are weight hyperparameters. The weights were set to $\tau=1$ and $\gamma=40$.

\subsection{Discussion} \label{subsection:comparisonSec}
Recently, attention-based semantic segmentation methods~\cite{Zhao2020CPSS-18,wang2018Nonlocal-5,wang2019AsyNonlocal-6,Fu2019DANet-7,ISNet2021ICCV} have achieved successful improvements, but they require too much complexity in the computation. 
These methods involve leveraging matrix multiplication to capture the correlation between all pixels and classes. Specifically, assuming the feature size to be $H\times W$, the attention weight matrix of these methods should have a size of $(H\cdot W)\times (H\cdot W)$. It should be noted that the segmentation results in better performance as the resolution of the input image increases. 
However, our CDGNet statistically calculates each class distribution in the horizontal and vertical directions. In contrast, to specifically consider the distribution properties of each class, the required distribution feature size is solely $H\times C+C\times W$. Note that the feature size ($(H+W)\times C$) of our method is much smaller than that of the attention weight matrix. 
The memory requirement decreases significantly in this case. 
When scaling the feature based on the distribution map, solely matrix element-wise multiplication and addition are required as shown in~\cref{fig:hwdnetwork}, which is also more computationally efficient than the matrix multiplication used in the attention-based method. Our CDGNet is, therefore, considerably lightweight compared to attention-based approaches.

Other related works that involve semantic tree~\cite{Ji2020LSNT-22} or human pose~\cite{Zhang2020CorrPM-8} require constructing different trees or human-labeling additional annotations for the various datasets, which constrains the generality of deployment. By comparison, our model does not need to adjust any parts of the architecture when applied to different datasets and can serve as a plugin module for any scene parsing network, as shown in the bottom part of ~\cref{fig:FusedGTImg}. 
Because most of the computational complexity occurs in 1D convolution, it is considerably lightweight and generic. 
Regarding the human pose-based method, despite the additional utilization of the pose estimator, the information obtained is only about the location of the head, body, arms, and legs, which are the pivotal human parts. Moreover, the final human parsing performance depends on the performance of the pose estimator.
Our algorithm utilizes the horizontal and vertical class distribution maps of each part of the human, which provides not only the statistical position of each part, but also the distribution range. Our method utilizes the distribution maps of all target classes including hat, sock, etc., which have not been included in the results of the pose estimator.
The non-local module has been utilized to capture the long-range contextual appearance information for each pixel based on self-attention methods~\cite{Fu2019DANet-7, Equivarian2020, Zhang2020CorrPM-8}. Our network can employ the non-local module, achieving higher performance. We conjecture that our class spatial distribution feature and the non-local feature are complementary to each other to provide better performance for human parsing.

\begin{table*}[t]
\centering
\setlength\tabcolsep{2.6pt}
\resizebox{\textwidth}{!}{%
\begin{tabular}{r c c c c c c c c c c c c c c c c c c c c c}
\toprule
Method 
&hat     &hair   &glove   &glass   &u-cloth &dress 
&coat    &sock   &pants   &j-suits &scarf   &skirt 
&face    &l-arm  &r-arm   &l-leg   &r-leg   &l-shoe 
&r-shoe  &bkg    &Avg    \\  
\midrule
Attention~\cite{Chen2016Attentiontoscale}    &58.87&66.78&23.32&19.48&63.20&29.63&49.70&35.23&66.04&24.73&12.84&20.41&70.58&50.17&54.03&38.35&37.70&26.20&27.09&84.00&42.92 \\
DeepLab~\cite{Chen2017Deeplab-1624}                      &56.48&65.33&29.98&19.67&62.44&30.33&51.03&40.51&69.00&22.38&11.29&20.56&70.11&49.25&52.88&42.37&35.78&33.81&32.89&84.53&44.03 \\
MMAN~\cite{LuoMMAN2018}                         &57.66&65.63&30.07&20.02&64.15&28.39&51.98&41.46&71.03&23.61&9.65&23.20&69.54&55.30&58.13&51.90&52.17&38.58&39.05&84.75&46.81 \\
SS-NAN~\cite{Zhao2017Self-SupervisedAgg}                       &63.86&70.12&30.63&23.92&70.27&33.51&56.75&40.18&72.19&27.68&16.98&26.41&75.33&55.24&58.93&44.01&41.87&29.15&32.64&88.67&47.92 \\
JPPNet~\cite{Liang2018JPPNet}                        &63.55&70.20&36.16&23.48&68.15&31.42&55.65&44.56&72.19&28.39&18.76&25.14&73.36&61.97&63.88&58.21&57.99&44.02&44.09&86.26&51.37 \\
CE2P~\cite{Ruan2019CE2P-21}                         &65.29&72.54&39.09&32.73&69.46&32.52&56.28&49.67&74.11&27.23&14.19&22.51&75.50&65.14&66.59&60.10&58.59&46.63&46.12&87.67&53.10 \\
SNT~\cite{Ji2020LSNT-22}                          &66.90&72.20&42.70&32.30&70.10&33.80&57.50&48.90&75.20&32.50&19.40&27.40&74.90&65.80&68.10&60.03&59.80&47.60&48.10&88.20&54.70\\
CorrPM~\cite{Zhang2020CorrPM-8}                       &66.20&71.56&41.06&31.09&70.20&37.74&\underline{57.95}&48.40&75.19&32.37&23.79&29.23&74.36&66.53&68.61&62.80&62.81&49.03&49.82&87.77&55.33\\
SCHP~\cite{li2020self}                       &69.96&73.55&\underline{50.46}&40.72&69.93&\underline{39.02}&57.45&54.27&76.01&32.88&26.29&31.68&76.19&68.65&70.92&67.28&66.56&55.76&56.50&88.36&58.62\\
\midrule
Ours                                &\underline{70.71}&\underline{74.43}&\textbf{50.68}&\underline{41.72}&\underline{70.92}&38.51&57.60&\underline{54.30}&\underline{77.65}&\underline{33.86}&\textbf{30.59}&\underline{31.98}&\underline{76.89}&\underline{70.93}&\underline{73.03}&\underline{70.30}&\underline{68.92}&\underline{58.17}&\underline{58.67}&\underline{88.83}&\underline{59.93} \\
Ours$^{\dagger}$                   
&\textbf{71.06}&\textbf{74.61}&50.13&\textbf{42.09}&\textbf{71.58}&\textbf{40.00}&\textbf{58.73}&\textbf{55.25}&\textbf{77.92}&\textbf{34.32}&\underline{30.05}&\textbf{32.97}&\textbf{77.12}&\textbf{71.25}&\textbf{73.35}&\textbf{70.54}&\textbf{69.26}&\textbf{58.24}&\textbf{58.75}&\textbf{88.86}&\textbf{60.30} \\
\bottomrule
\end{tabular}}
\vspace{-0.2cm}
\caption{
Per-class intersection over union (IoU) comparison of the validation set of LIP. Here, $^{\dagger}$ is the parsing result averaged over the multiscaled image pyramids with flipping in the inference time. The result of SCHP~\cite{li2020self} is made by a single image in the inference time. The best values are marked in bold, and the second-best values are marked with an underline. 
}
\label{tab:perclassIoU}
\vspace{-0.4cm}
\end{table*}

\section{Experiments}

\subsection{Datasets}
\textbf{Look Into Person (\textbf{LIP}):}
The LIP dataset~\cite{Gong2017LIP} was utilized in the LIP Challenge 2016 for human parsing tasks. In total, 50,462 images were provided, including 30,462 images for training and 10,000 for validation. These images are finely labeled at the pixel level with 19 semantic human part classes (including 6 body parts and 13 items of clothing) and one background category.

\textbf{Active Template Regression (\textbf{ATR}):}
The ATR dataset~\cite{Liang2015CCNNetwork} contains 18 semantic category labels (including face, sunglasses, hat, scarf, hair, upper clothes, left arm, right arm, belt, pants, left leg, right leg, skirt, left shoe, right shoe, bag, dress, and background). In total, 17,700 images were included in ATR. Following\cite{Zhang2020CorrPM-8}, 16,000 images were utilized for training, 1,000 for testing, and 700 for validation.

\textbf{Crowd Instance-level Human Parsing (\textbf{CIHP}):}
The CIHP dataset~\cite{Gong2018PGN} is a large-scale dataset that provides 38,280 multi-person images with 20 semantic parts, including the background. The images in the CIHP were collected from a real-world scenario. Persons in the images appear with challenging poses and viewpoints, heavy occlusions, and show in a wide range of resolutions~\cite{Ke2019CIHP}. The dataset is elaborately annotated to benefit the semantic understanding of multiple people in a real situation. It comprises 28,280 training, 5,000 validation and 5,000 test images.

\subsection{Experimental Settings}
\textbf{Implementation Details:}
We employed the basic structure and network settings of CE2P~\cite{Ruan2019CE2P-21} as the baseline. 
The feature from the backbone with 512 channels feeds into two parallel 1D spatial pooling along the height or width axes. The channel number of the horizontal and vertical features is reduced to 256 through a 1D convolution with a kernel size of $3$. 
Subsequently, we generate the predicted class distribution map, $P_h$ of size $H\times N$ and $P_v$ of size $N\times W$, using the convolution with kernel size 3 and sigmoid, which is guided by the proposed CDG loss. 
Meanwhile, with a 1D convolution of kernel size $7$ and $sigmoid$ activation, the channel numbers of the horizontal and vertical features are increased to 512. Subsequently, the guided features are upsampled using the bilinear interpolation to $H\times W \times 512$, and they are merged into the spatial guidance map by element-wise summation. We perform matrix element-wise multiplication between the spatial guidance map and the input feature. We finally reduce its channel number to 512 using $3\times 3 \ convolution+BN+ReLU$ to extract the CDG feature.

\textbf{Data Augmentation:}
In the training procedure, we adopt the mean subtraction, random scaling in the range of $[0.5, 1.25]$, random color jittering, and random left-right flipping as the basic data augmentation. We randomly crop the large image or pad the small images into a fixed size for training (i.e., $473\times 473$ for LIP and CHIP, $512 \times 512$ for ATR).

\textbf{Training:}
We adopt ResNet-101~\cite{He2016MaskRCNN-26} pre-trained using the ImageNet database~\cite{krizhevsky2012imagenet} as the backbone. After appending the CDG module in the context embedding module of CE2P~\cite{Ruan2019CE2P-21}, the network is trained for $150$ epochs on the LIP and CIHP datasets, and $250$ epochs on the ATR dataset. Stochastic gradient descent (SGD) is utilized as the optimizer, and the initial learning rate is set to $3e^{-3}$. We adopt the polynomial learning rate strategy $r=(1-\frac{cur\_iter}{total\_iter})^{0.9}$, where $cur\_iter$ and $total\_iter$ represent the current iteration and total iteration numbers, respectively. The momentum is set to $0.9$, and the weight decay is $5e^{-4}$. We adopt cross-entropy loss when training on all datasets.

\textbf{Inference:}
In the inference time, the pixel accuracy (pixAcc), mean accuracy, and mean pixel Intersection-over-union (mIoU) are leveraged as the evaluation metrics for the LIP dataset, pixel accuracy, precision, recall, and F-1 scores for the ATR dataset, and mIoU for the CIHP dataset. 
Similar to~\cite{XiaoPCNet2020, Ke2019CIHP, li2020self,Wang2020HTypePRR-20}, we averaged the results using 3-scale image pyramids of different scales $[0.75, 1.0, 1.25]$ and flipping for further performance improvement.

\subsection{Quantitative and Qualitative Experiments}
\label{sec:result}

To achieve the best performance in human parsing, in this paper, we make use of mIOU loss~\cite{li2020self,Wang2020HTypePRR-20} and non-local module~\cite{Equivarian2020} with our proposed method. 
Quantitative experiments were performed by comparison with well-known human parsing methods using the LIP, ATR, and CIHP datasets. Through these experimental results, we verified that our method can achieve meaningful and precise performance regardless of the difficulty of the human image and number of persons. 

\textbf{Performance on LIP database:}
Table~\ref{tab:perclassIoU} presents the performance comparison of the proposed method to other methods on the LIP validation set. Basically, our method achieves the best performances compared with others as well as the baseline method through all 20 classes of LIP dataset. 
What we should note here is that, compared with others, we significantly improve the IoUs of small parts of human parsing, for example, glasses, scarves, and shoes. In this respect, we can infer that our positional distributions using all classes are better than the positions of the selected human parts estimated by the complex pose estimator of CorrPM~\cite{Zhang2020CorrPM-8}. From the viewpoint of the IoU of the scarf, our method achieves approximately 6\% better than CorrPM~\cite{Zhang2020CorrPM-8} and 4\% compared to SCHP~\cite{li2020self}. 
It is mainly because the position of the scarf could be determined around the head and upper body, and we naturally have a higher chance to teach the network using the supervision signals for the horizontal and vertical class distributions to possess the corresponding information. 
Consequently, our CDGNet exhibits the state-of-the-art performance (60.30\% mIoU) on LIP, and our result is approximately 1\% better than the human part relation-based HHP~\cite{Wang2020HTypePRR-20} (59.25\% mIoU) and the label self correction-based SCHP~\cite{li2020self} (59.36\% mIoU), as summarized in Table~\ref{tab:example1}. 
When compared with the baseline (CE2P), our method achieves $7.20\%$ higher in terms of mIoU and $1.5\%$ and $8.2\%$ higher in terms of pixel accuracy and mean accuracy, respectively. We conclude from these results that our method, which considers the class distributions correlated with the whole class, can yield more promising results than the others using a few relations of the selected human parts in human paring. 

\begin{table}[t]\centering
\resizebox{0.9\linewidth}{!}{%
\small
\begin{tabularx}{\linewidth}{ r c c c}
\toprule
Method    & Pixel Acc. & Mean Acc. & mIoU  \\ 
\midrule
CE2P~\cite{Ruan2019CE2P-21}                       & 87.37     & 63.20    &  53.10\\
SNT~\cite{Ji2020LSNT-22}                        & 88.05     & 66.42    &  54.73\\
Double Attention~\cite{Fu2019DANet-7}    & -         & -        & 55.12\\
CorrPM~\cite{Zhang2020CorrPM-8}                     & 87.68     &67.21    &  55.33\\
BGNet~\cite{Zhang2020BGNet-23}&-&-&56.82\\
ISNet~\cite{ISNet2021ICCV} &-&-&56.96\\
MCIBISS~\cite{MCIBISS2021ICCV} &-&-&56.99\\
PCNet~\cite{XiaoPCNet2020} & - & - &57.03\\
HHP~\cite{Wang2020HTypePRR-20} &\textbf{89.05}& \underline{70.58}&59.25 \\
SCHP~\cite{li2020self} &-&-&\underline{59.36}\\
\midrule
Ours & \underline{88.86}     &\textbf{71.49}    &  \textbf{60.30}\\
\bottomrule
\end{tabularx}%
}
\vspace{-0.2cm}
\caption{Comparison of different methods on the validation set of the LIP dataset. 
The bold is the best one and the underline is the second best. 
}
\label{tab:example1}
\vspace{-0.4cm}
\end{table}

\textbf{Performance on ATR database:}
The ATR database comprises a relatively small number of images compared with the LIP dataset, and most of the images are captured with a single person's frontal pose.  
As presented in Table~\ref{tab:example2}, our method achieves the best performance in terms of pixel accuracy and the second rank for foreground pixel accuracy, precision, recall, and F-1 measures. 
Specifically, compared with CorrPM~\cite{Zhang2020CorrPM-8}, our method showed better results, except for foreground pixel accuracy, and precision. However, note that the performance gap between our method and CorrPM in the LIP database is approximately 5\% at mIoU. When comparing our method with HHP~\cite{Wang2020HTypePRR-20}, our method achieves better results in terms of pixel accuracy, foreground accuracy, and precision, but lower results at recall and F-1. We note that our method performs 1\% better on the LIP dataset. 

\begin{table}[t]\centering\small
\resizebox{0.9\linewidth}{!}{%
\begin{tabularx}{\linewidth}{r c c c c c}
\toprule
Method &Acc.&F.G.Acc.&Pre.&Recall&F-1 \\ 
\midrule
DeepLabV2~\cite{Chen2017Deeplab-1624} &94.42 &82.93&69.24 & 78.48&73.53\\
Attention~\cite{Chen2016Attentiontoscale} & 95.41&85.71& 81.30& 73.55&77.23\\
CoCNN~\cite{Liang2015CCNNetwork} & 96.02&83.57&84.59& 77.66&80.14\\
TGPNet~\cite{LuoTGPNet2018}  & 96.45&87.91&83.36 & 80.22&81.76\\
CorrPM~\cite{Zhang2020CorrPM-8}  &\underline{97.12}&\textbf{90.40}& \textbf{89.18}&83.93&86.12\\
HHP~\cite{Wang2020HTypePRR-20}  &96.87&89.23&86.17&\textbf{88.35}&\textbf{87.25}\\
\midrule
Ours & \textbf{97.39} & \underline{90.19} & \underline{87.46} & \underline{86.87} & \underline{87.16}\\
\bottomrule
\end{tabularx}%
}
\vspace{-0.2cm}
\caption{
Comparison of accuracy, foreground accuracy, precision, recall and F-1 scores on the ATR test set. The bold font is the best value, and the underline is the second. }
\label{tab:example2}
\vspace{-0.2cm}
\end{table}

\begin{table}[t]\centering\normalsize
\resizebox{0.9\linewidth}{!}{%
\setlength{\tabcolsep}{0.5cm}
\begin{tabularx}{\linewidth}{rcc}
\toprule
Method&Backbone 
&mIoU\\ 
\midrule
PGN~\cite{Gong2018PGN} & DeepLabV2
&55.80\\
Graphonomy~\cite{Ke2019CIHP}  & DeepLabV3+
&58.58\\
M-CE2P~\cite{Ruan2019CE2P-21} &ResNet101 &59.50\\ 
CorrPM~\cite{Zhang2020CorrPM-8}  &ResNet101 &60.18\\
SNT~\cite{Ji2020LSNT-22} &ResNet101& 
60.87\\
PCNet~\cite{XiaoPCNet2020} &ResNet101 
&\underline{61.05}\\
\midrule
Ours   &ResNet101 
&\textbf{65.56}\\
\bottomrule
\end{tabularx}%
}
\vspace{-0.2cm}
\caption{
Performance comparison with state-of-the-art method on CIHP dataset. The bold font is the best value, and the underline is the second. }
\vspace{-0.4cm}
\label{tab:CIHP}
\end{table}

\textbf{Performance on CIHP database:}
In the multi-person-based database CIHP, our CDGNet achieves the best performance of $65.56\%$ among the previous works based on ResNet-101, and outperforms the baseline and PCNet by $6.1\%$ and $4.5\%$, respectively, as presented in Table~\ref{tab:CIHP}. 
Note that improving the multiple human parsing task is highly challenging, but we do not design additional methods for multiple person-based human parsing and obtain $5.5\%$ higher performance than CorrPM based on the pose information. 
Moreover, the input image size is $473\times473$, which is much smaller than that of PCNet, which leverages an image size of $512\times512$. In the CIHP, we are required to train a more general human parsing model for the case of multiple persons. For example, when multiple persons stand close together, occlusion frequently occurs, which does not occur in a single person. Our method can overcome this problem because the horizontal class distribution is statistically made by accumulating the features of multiple persons existed in an image. In this respect, the noise of one or two persons among several persons does not significantly affect the overall performance of the proposed class distributions.  


\subsection{Qualitative Results}
\begin{figure}[t]
\centering
\includegraphics[width=0.9 \linewidth]{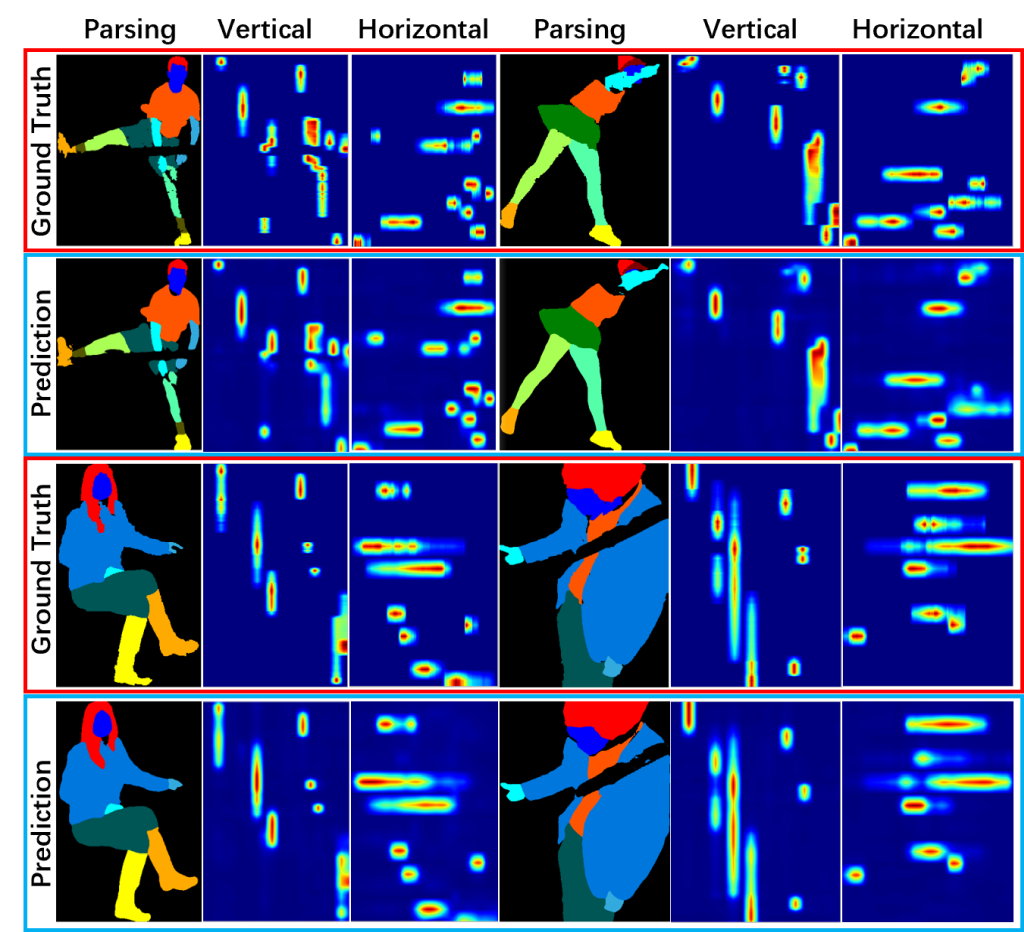}
\vspace{-0.2cm}
\caption{Visualization of the predicted class distribution map.
CDGNet generates the horizontal and vertical distributions of the human parts under the supervision of corresponding distribution labels.
Deeper color denotes higher probability of class existence. }
\vspace{-0.4cm}
\label{fig:horVal}
\end{figure}

\begin{figure}[t]
\centering
\includegraphics[width=0.95 \linewidth]{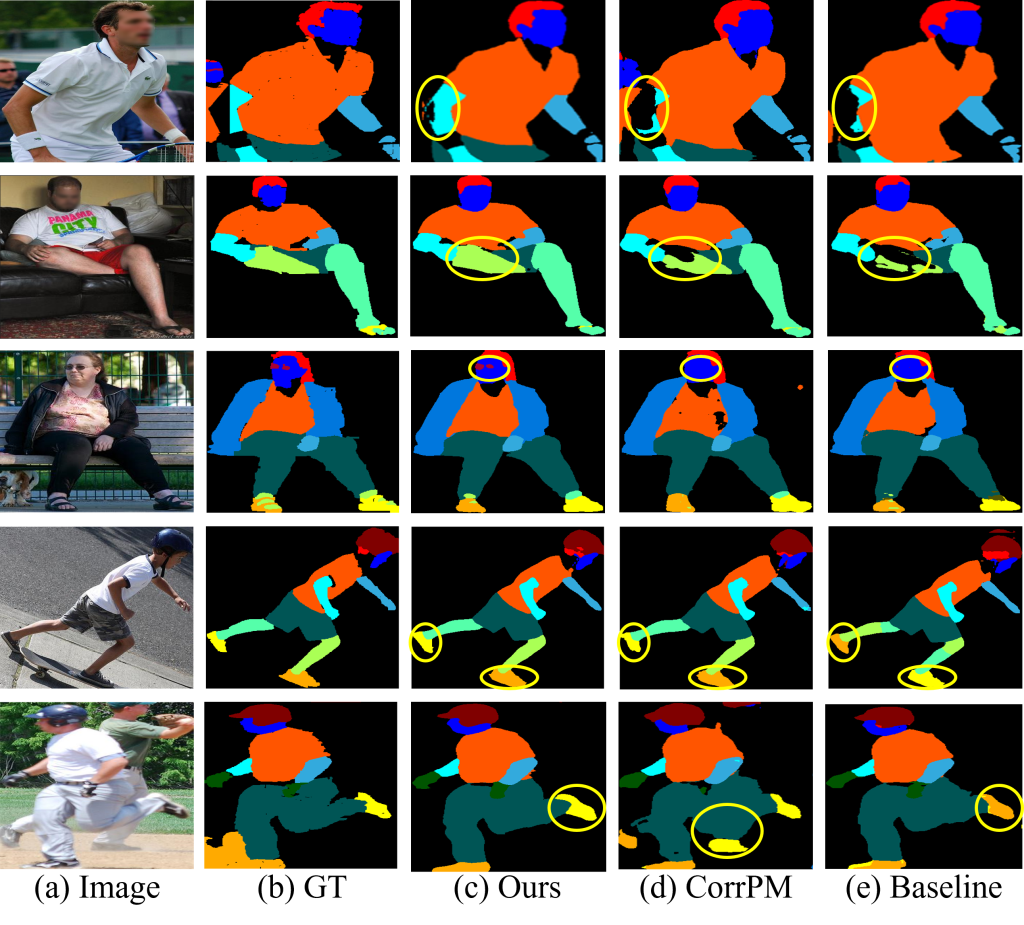}
\vspace{-0.2cm}
\caption{Visualization examples of different methods on the LIP validation dataset. We point out the differences between ours and others with yellow ellipses. To avoid the negative societal impact, when showing faces in this paper, face regions are blurred.}
\vspace{-0.4cm}
\label{fig:resultLIP}
\end{figure}

In this section, we first verify that the proposed network correctly predicts the horizontal and vertical class distributions. As shown in \cref{fig:horVal}, the predicted results are very similar to Ground Truth (GT) generated from the original human parsing label. Therefore, we can infer that the horizontal and vertical labels guided the proposed method well during the training procedure, and horizontal and vertical class distributions can reflect individually where each class exists with high probability. \cref{fig:resultLIP} validates that our method achieves improved performance on various human parts with manifest distribution characteristics, such as the right arm in the first row, right leg in the second row, and glasses in the third row. 
In the fourth row examples of \cref{fig:resultLIP}, our method and CorrPM indicate the correct positions of the left and right shoes while the baseline does not; thus, we can infer that the proposed class distribution guided network can distinguish the left and right parts of human parsing successfully. In the last row example, the baseline makes other errors on the left and right shoes, and CorrPM indicates the poor result owing to the occlusion problem. However, our method correctly predicts the human parsing result.

\begin{table}[t]\centering\small
\resizebox{0.9\linewidth}{!}{%
\begin{tabular}{c c c c|c c c } 
\toprule
\multicolumn{4}{c|}{Method} & \multicolumn{1}{c}{\multirow{2}{*}{Pixel Acc.}} & \multirow{2}{*}{Mean Acc.} & \multirow{2}{*}{mIoU} \\ 
B   & G   & I   & N  & \multicolumn{1}{c}{}   &  &      \\ \midrule
\checkmark&-&-&-&87.37&63.20&53.10\\ 
\checkmark&\checkmark&-&-&88.41&68.87&57.72  \\ 
\checkmark&\checkmark&\checkmark&-&88.53&70.89&59.04\\
\checkmark&\checkmark&\checkmark&\checkmark&88.65&71.98&59.93\\ \bottomrule
\end{tabular}
}
\vspace{-0.2cm}
\caption{
Each component of our method is evaluated on the LIP validation set, including baseline (B), CDG module (G), IOU loss (I), non-local module (N).}
\label{tab:Ablation}
\vspace{-0.4cm}
\end{table}

\subsection{Ablation Studies} 
As Table~\ref{tab:Ablation} presents, when we soley integrate the CDG module with the baseline, it makes a dramatic improvement in terms of mIoU, from $53.1\%$ to $57.72\%$. 
Note that this performance is significantly higher than that of attention-based and context-based methods, the results of which are presented in Table~\ref{tab:example1}. 
For instance, it is $2.5\%$ higher than that of the double attention method~\cite{Fu2019DANet-7} and $0.72\%$ higher than the latest image- and semantic-level context method~\cite{ISNet2021ICCV}.
By further adopting mIoU loss that has been commonly utilized in the latest works~\cite{Wang2020HTypePRR-20, li2020self} to achieve the state-of-the-art accuracy, we achieved a higher performance of $59.04\%$. Finally, we verify that our CDGNet can learn complementary features to the non-local module and unify the non-local module with CDGNet, which results in the performance improvement from $59.04\%$ to $59.93\%$ in terms of mIoU. 


\section{Conclusion}
In this paper, we proposed a effective and efficient method, called CDGNet, for human parsing. We exploited the pixel labeling of each category to produce the horizontal and vertical class distributions to reflect the distribution rule of each human part. With the guidance of the generated class distributions of each class in the horizontal and vertical directions,
the network had more chance of learning the distribution consistent with the structural prior of the human body.
It significantly benefited the labeling of each pixel in images with only one person and multiple persons as well. 
Extensive quantitative and qualitative comparisons indicated that the proposed CDGNet perform favorably against the state-of-the-art human parsing approaches. 

\noindent\textbf{Limitation}: We assume that the number of classes is mostly about twenty (e.g., LIP, ATR, CIHP) and CDGNet can make the best accuracy. However, we note that CDGNet may not perform the best performance when using the small size of classes (e.g., one or two classes) in human parsing.

\noindent\textbf{Potential Negative Societal Impact}: CDGNet is a generic technology with many potential applications for human images. We are unaware of all potential works but can imagine that each application has its own merits and societal impacts depending on the intentions of the users. 

\noindent\textbf{Acknowledgment}: This work is partially supported by IITP grant/MSIT (No.2021-0-00951, Development of Cloud based Autonomous Driving AI learning Software), ITRC grant (IITP-2021-2018-0-01431), BK21 Four (NRF5199991014091), IITP grant/MSIT (AI Innovation Hub, 2021-0-02068), Korea, and Tianjin Science and Technology Program (No. 19PTZWHZ00020), China.

{\small

}

\end{document}